\documentclass[sigconf]{acmart}

\AtBeginDocument{%
  \providecommand\BibTeX{{%
    \normalfont B\kern-0.5em{\scshape i\kern-0.25em b}\kern-0.8em\TeX}}}

\copyrightyear{2021}
\acmYear{2021}
\setcopyright{acmlicensed}\acmConference[SIGIR '21]{Proceedings of the 44th International ACM SIGIR Conference on Research and Development in Information Retrieval}{July 11--15, 2021}{Virtual Event, Canada}
\acmBooktitle{Proceedings of the 44th International ACM SIGIR Conference on Research and Development in Information Retrieval (SIGIR '21), July 11--15, 2021, Virtual Event, Canada}
\acmPrice{15.00}
\acmDOI{10.1145/3404835.3463004}
\acmISBN{978-1-4503-8037-9/21/07}

\usepackage{graphicx} 
\usepackage{latexsym}
\usepackage{url}
\usepackage{booktabs}       
\usepackage{amsfonts}       
\usepackage{nicefrac}       
\usepackage{microtype}      
\usepackage{algorithmic}
\usepackage{algorithm}
\usepackage{wrapfig}
\usepackage{lipsum}
\usepackage{xspace}
\usepackage{bm}
\usepackage{subfigure} 
\usepackage[skip=0pt]{caption}
\usepackage{paralist}
\usepackage{float}
\usepackage{tabularx}
\usepackage{multirow}
\usepackage{diagbox}
\usepackage{tikz}
\usepackage{array, makecell} 
\usepackage{amsmath}
\usepackage[inline]{enumitem}
\usepackage{acronym}
\usepackage{bm}

 \newcommand\BibTeX{B{\sc ib}\TeX}

\newcounter{todocnt}

\acrodef{RL}{Reinforcement Learning}
\acrodef{DRL}{Deep Reinforcement Learning}
\acrodef{IRL}{Inverse Reinforcement Learning}
\acrodef{SERP}{search engine result page}
\acrodef{IR}{Information Retrieval}
\acrodef{MDP}{Markov Decision Process}
\acrodef{MaxEnt-IRL}{Maximum Entropy Inverse Reinforcement Learning}
\acrodef{DM-IRL}{Distance Minimization Inverse Reinforcement Learning}
\acrodef{MMI}{Maximum Mutual Information} 
\acrodef{DNN}{Deep Neural Networks}
\acrodef{RNN}{Recurrent Neural Networks}
\acrodef{MLP}{Multilayer Perceptron}
\acrodef{GRU}{Gated Recurrent Net}

\begin{document}
\fancyhead{}
\title[Improving Response Quality with Backward Reasoning]{Improving Response Quality with Backward Reasoning\\ in Open-domain Dialogue Systems}

\author{Ziming Li}
\orcid{}
\email{z.li@uva.nl}
\affiliation{%
\institution{University of Amsterdam}
\city{Amsterdam}
\country{The Netherlands}
}

\author{Julia Kiseleva}
\orcid{}
\email{julia.kiseleva@microsoft.com}
\affiliation{%
\institution{Microsoft}
\city{Redmond}
\country{United States}
}

\author{Maarten de Rijke}
\orcid{0000-0002-1086-0202}
\email{m.derijke@uva.nl}
\affiliation{%
\institution{University of Amsterdam \& Ahold Delhaize}
\city{Amsterdam}
\country{The Netherlands}
}

\begin{abstract}
Being able to generate informative and coherent dialogue responses is crucial when designing human-like open-domain dialogue systems. Encoder-decoder-based dialogue models tend to produce generic and dull responses during the decoding step because the most predictable response is likely to be a non-informative response instead of the most suitable one. 
To alleviate this problem, we propose to train the generation model in a bidirectional manner by adding a backward reasoning step to the vanilla encoder-decoder training. The proposed backward reasoning step pushes the model to produce more informative and coherent content because the forward generation step's output is used to infer the dialogue context in the backward direction. The advantage of our method is that the forward generation and backward reasoning steps are trained simultaneously through the use of a latent variable to facilitate bidirectional optimization. Our method can improve response quality without introducing side information (e.g., a pre-trained topic model). 
The proposed bidirectional response generation method achieves state-of-the-art performance for response quality. 
\end{abstract}

\ccsdesc[500]{Information systems~Chat}
\ccsdesc[300]{Information systems~Question answering}

\keywords{Open-domain dialogue system; response generation}

\maketitle

\section{Introduction}
\label{sec:introduction}
\begin{figure}[t]
\centering
   \includegraphics[clip, width=0.8\columnwidth]{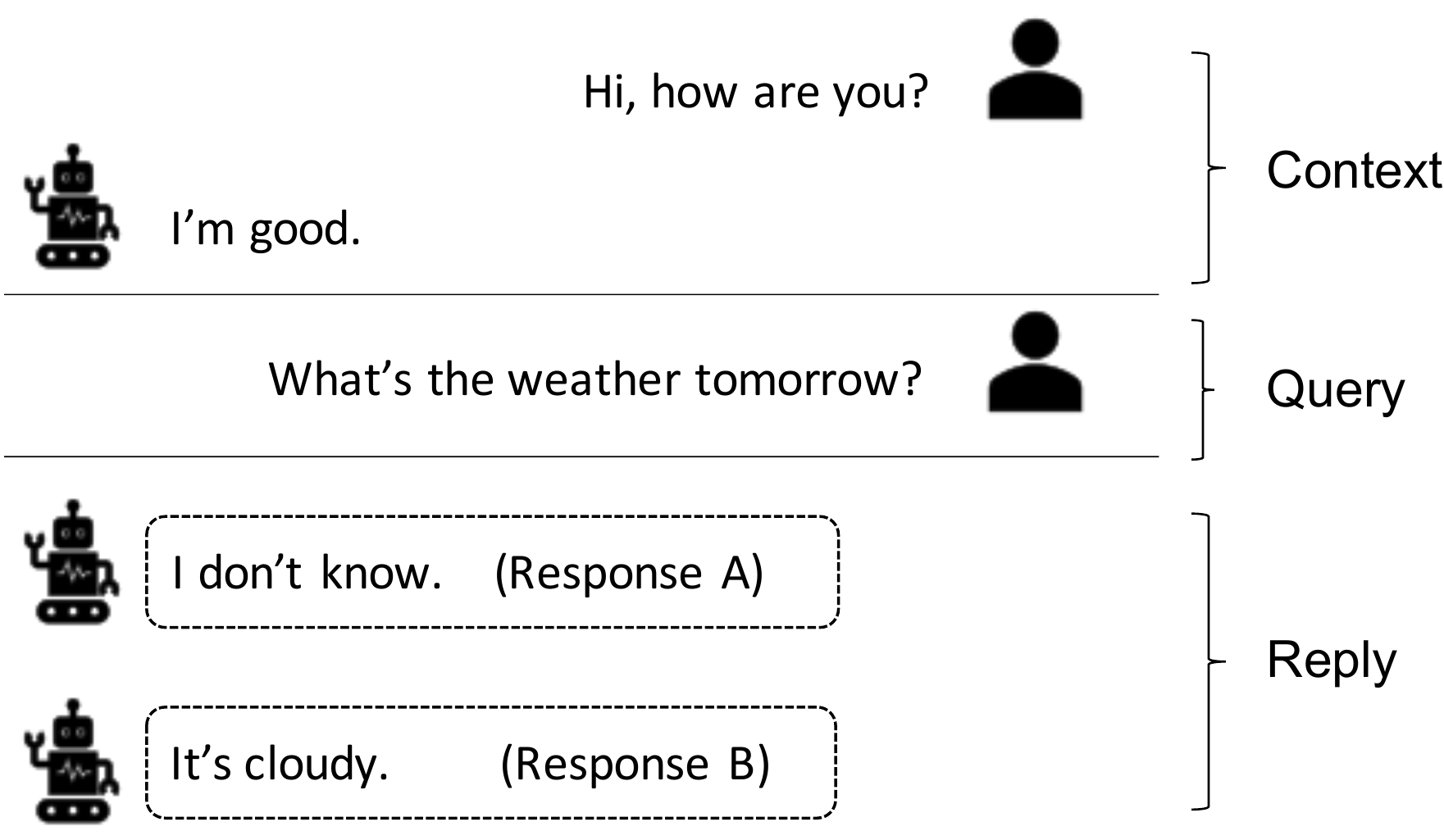}
   \caption{A more informative response (Response B in the figure) can provide information that helps to infer the query content given the dialogue context.}
   \vspace{-0.5 cm}
   \label{fig:back-reasoning}
\end{figure}

Recently developed end-to-end dialogue systems are trained using large volumes of human-human dialogues to capture  underlying interaction patterns~\citep{li2015diversity,li2017adversarial,xing2017topic,khatri2018advancing,vinyals2015neural,zhang2019dialogpt,bao2019plato}. A commonly used approach to designing data-driven dialogue systems is to use an encoder-decoder framework: feed the dialogue context to the encoder, and let the decoder output an appropriate response. Building on this foundation, different directions have been explored to design dialogue systems that tend to interact with humans in a coherent and engaging manner~\citep{li2016deep,li2019dialogue,wiseman2016sequence,baheti2018generating,xing2016topic,zhang2018personalizing}. However, despite significant advances, there is still room for improvement in the quality of machine-generated responses.

An important problem with encoder-decoder dialogue models is their tendency to generate generic and dull responses, such as \emph{``I don't know''} or \emph{``I'm not sure''}~\citep{li2015diversity,baheti2018generating,li2016deep,jiang-why-2018}. 
There are two types of methods for dealing with this problem.
The first introduces updating signals during training, such as modeling future rewards (e.g., ease of answering) by applying reinforcement learning~\citep{li2016deep,li2019dialogue}, or bringing variants or adding constraints to the decoding step~\citep{wiseman2016sequence,li2015diversity,baheti2018generating}. 
The second type holds that, by itself, the dialogue history is not enough for generating high-quality responses, and side information should be taken into account, such as topic information~\citep{xing2016topic,xing2017topic} or personal user profiles~\citep{zhang2018personalizing}. Solutions relying on large pre-trained language models, such as DialoGPT~\citep{zhang2019dialogpt}, can be classified into the second family as well.

In this paper, we propose to train dialogue generation models \emph{bidirectionally} by adding a backward reasoning step to the vanilla encoder-decoder training process. 
We assume that the information flow in a conversation should be coherent and topic-relevant. Given the dialogue history, neighboring turns are supposed to have a tight topical connection to infer the partial content of one turn given the previous turn \emph{and vice versa}. 
Inferring the next turn given the (previous) conversation history and the current turn is the traditional take on the dialogue generation task. 
We extend it by adding one more step: given the dialogue history and the next turn, we aim to infer the content of the current turn. We call the latter step \emph{backward reasoning}. We hypothesize that this can push the generated response to be more informative and coherent: it is unlikely to infer the dialogue topic given a generic and dull response in the backward direction. 
An example is shown in Figure~\ref{fig:back-reasoning}. Given the dialogue context and \emph{query},\footnote{We use \emph{query} to distinguish the current dialogue turn from the context and the response; \emph{query} is not necessarily a real query or question as considered in search or question-answering tasks.} we can predict the reply following a traditional encoder-decoder dialogue generation setup. In contrast, we can infer the content of \emph{query} given the context and reply as long as the reply is informative. 
Inspired by~\citet{zheng2019mirror}, we introduce a latent space as a bridge to simultaneously train the encoder-decoder model from two directions. 
Our experiments demonstrate that the resulting dialogue generation model, called \emph{Mirror}, benefits from this bidirectional training process.

Overall, our work provides the following contributions:
\begin{enumerate}[leftmargin=*,label=\textbf{C\arabic*},nosep]

\item We introduce a dialogue generation model, \emph{Mirror}, for generating high quality responses in open-domain dialogue systems;

\item We define a new way to train dialogue generation models bidirectionally by introducing a latent variable; and

\item We obtain improvements in terms of dialogue generation performance with respect to human evaluation on two datasets.
\end{enumerate}

\section{Related Work}
Conversational scenarios being considered today are increasingly complex, going beyond the ability of rule-based dialogue systems~\citep{weizenbaum1966eliza}. \citet{ritter2011data} propose a data-driven approach to generate responses, building on phrase-based statistical machine translation. Neural network-based models have been studied to generate more informative and interesting responses~\citep{sordoni2015neural,vinyals2015neural,serban2016hred}.
\citet{serban2017hierarchical} introduce latent stochastic variables that span a variable number of time steps to facilitate the generation of long outputs. 
Deep reinforcement learning methods have also been applied to generate coherent and interesting responses by modeling the future influence of generated responses \citep{li2016deep,li2019dialogue}.  Retrieval-based methods are also popular in building dialogue systems by learning a matching model between the context and pre-defined response candidates for response selection~\citep{qiu2020if,tao2019multi,wu2016sequential,gu2020speaker}. Our work focuses on response \emph{generation} rather than \emph{selection}.

Since encoder-decoder models tend to generate generic and dull responses, \citet{li2015diversity} propose using maximum mutual information as the objective function in neural models to generate more diverse responses.
\citet{xing2017topic} consider incorporating topic information into the encoder-decoder framework to generate informative and interesting responses. 
To address the dull-response problem, \citet{baheti2018generating} propose incorporating side information in the form of distributional constraints over the generated responses. \citet{su2020diversifying} propose a new perspective to diversify dialogue generation by leveraging non-conversational text. Recently, pre-trained language models, such as GPT-2~\citep{radford2019language}, Bert~\citep{devlin2018bert}, XL-Net~\citep{yang2019xlnet}, have been proved effective for a wide range of natural language processing tasks. Several authors make use of pre-trained transformers to attain performance close to humans both in terms of automatic and human evaluation~\citep{zhang2019dialogpt,wolf2019transfertransfo,golovanov2019large}. Though pre-trained language models can perform well for general dialogue generation, they may become less effective without enough data or resources to support these models' pre-training. In this work, we show the value of developing dialogue generation models with limited data and resources.

The key distinction compared to previous efforts~\citep{li2015diversity,baheti2018generating} is our work is the first to use the original training dataset through a differentiable backward reasoning step, without external information.

\section{Method: \emph{Mirror}}
\label{sec:method}
\subsection{Problem setting}
In many conversational scenarios, the dialogue context is relatively long and contains a lot of information, while the reply (\emph{Response}) is short (and from a different speaker). This makes it difficult to predict the information in the context by only relying on the response in the backward direction. 
Therefore, we decompose the dialogue context into two different segments: the context $c$ and query $x$ (Figure~\ref{fig:back-reasoning}). 
Assuming that we are predicting the response at turn $t$ in a dialogue, the context $c$ will consist of the dialogue turns from $t-m$ to $t-2$ and the query $x$ corresponds to turn $t-1$. 
Here, we use the term \emph{query} to distinguish the dialogue turn at time step $t-1$ from the context $c$ and response $y$; as explained before, the term \emph{query} should not be confused with a query or question as in search or question-answering tasks. 
The value $m$ indicates how many dialogue turns we keep in the context $c$. 
We use $c_{all}$ to represent the concatenation of $c$ and $x$, which is also the original context before being decomposed. 
Our final goal is to predict the response $y$ given dialogue context $c$ and query $x$. 

\subsection{Mirror-generative dialogue generation}
\label{section:mirror-method}
\citet{shen2017conditional} propose to maximize the conditional log likelihood of generating response $y$ given context $c_{all}$, $\log p(y \mid c_{all})$,  and they introduce a latent variable $z$ to group different valid responses according to the context $c_{all}$. 
The lower bound of $\log p(y \mid c_{all})$ is given as:
\begin{equation}
\begin{split}
\log p(y \mid c_{all}) \geq{} &  \mathbb{E}_{z \sim q_\phi(z \mid c_{all}, y)} \log p_\theta (y \mid c_{all}, z)  - {}\\
& D_\mathit{KL}(q_\phi(z \mid c_{all}, y) \| p_\theta(z \mid c_{all})).
\end{split}
\label{eq:loss-vhred}
\end{equation}
In Eq.~\ref{eq:loss-vhred}, $q_\phi(z \mid c_{all}, y)$ is the posterior network while $p_\theta(z \mid c_{all})$ is the prior one.

Instead of maximizing the conditional log likelihood $\log p(y \mid c_{all})$, we propose to maximize $\log p(x,y \mid c)$, representing the conditional likelihood that $\langle x,y\rangle$ appears together given dialogue context $c$. 
The main assumption underlying this change is that in a conversation, the information flow between neighboring turns should be coherent and relevant, and this connection should be bidirectional. 
For example, it is not possible to infer what the query is about when a generic and non-informative reply ``I don't know'' is given as shown in Figure~\ref{fig:back-reasoning}. 
By taking into account the information flow from two different directions, we hypothesize that we can build a closer connection between the response and the dialogue history and generate more coherent and informative responses. Therefore, we propose to optimize $\log p(x,y \mid c)$ instead of $\log p(y\mid c_{all})$. 

Following \cite{kingma2013auto,shen2017conditional}, we choose to maximize the variational lower bound of $\log p(x,y \mid c)$, which is given as:
\begin{equation}
\begin{split}
\log p(x,y \mid c) \geq {} & \mathbb{E}_{z \sim q_\phi(z \mid c, x, y)} \log p_\theta (x, y \mid c, z)  - {}\\
& D_\mathit{KL}(q_\phi(z \mid c, x, y) \| p_\theta(z \mid c)),
\end{split}
\label{eq:loss-lb}
\end{equation}
where $z$ is a shared latent variable between context $c$, query $x$ and response $y$.  
Next, we explain how we optimize a dialogue system by maximizing the lower bound shown in  Eq.~\ref{eq:loss-lb} from two directions.

\begin{figure}[t]
\centering
   \includegraphics[clip, width=0.75\linewidth]{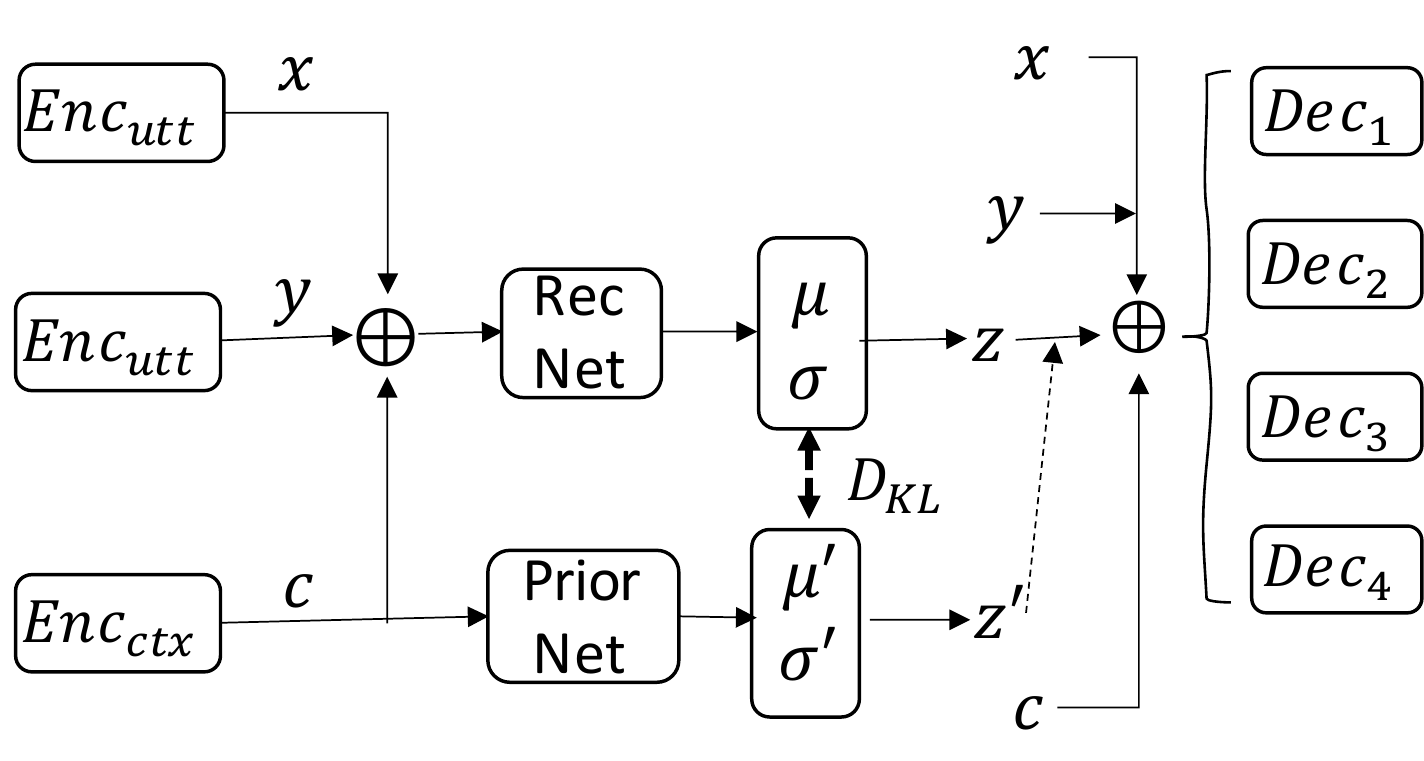}
   \caption{The main architecture of our model, \emph{Mirror}. It consists of three steps: information encoding, latent variable generation, and target decoding. }
   \vspace*{-0.5\baselineskip}
   \label{fig:framework}
\end{figure}

\subsubsection{Forward generation in dialogue generation}
With respect to the forward dialogue generation, we interpret the conditional likelihood $\log p_\theta (x, y \mid c, z)$ in the forward direction:
\begin{equation}
\begin{split}
\mbox{}\hspace*{-3mm}
\log p_\theta (x, y \mid c, z) 
=  \log p_\theta (y \mid c, z, x) + \log p_\theta (x \mid c, z).
\end{split}
\hspace*{-2mm}\mbox{}
\label{eq:cond-backward}
\end{equation}
Therefore, we can rewrite Eq.~\ref{eq:loss-lb} in the forward direction as:
\begin{equation}
\begin{split}
\log\, & p(x,y \mid c)\\
\geq{} & \mathbb{E}_{z \sim q_\phi(z \mid c, x, y)} [\log p_\theta (y \mid c, x, z) + \log p_\theta (x \mid c, z)] {}\\& - D_\mathit{KL}(q_\phi(z \mid c, x, y) \| p_\theta(z \mid c)).
\end{split}
\label{eq:forward-loss}
\end{equation}
We introduce $q_\phi(z \mid c, x, y)$ as the posterior network, also referred to as the recognition net, and $p_\theta(z \mid c)$ as the prior network.

\subsubsection{Backward reasoning in dialogue generation}
As in the forward direction, if we decompose the conditional likelihood $\log p_\theta (x, y \mid c, z)$ in the backward direction, we can rewrite Eq.~\ref{eq:loss-lb} as:
\begin{equation}
\begin{split}
\log\,& p(x,y \mid c) \\
\geq {} & \mathbb{E}_{z \sim q_\phi(z \mid c, x, y)} [\log p_\theta (x \mid c, y, z) + \log p_\theta (y \mid c, z)]  {}\\
&- D_\mathit{KL}(q_\phi(z \mid c, x, y) \| p_\theta(z \mid c)).
\end{split}
\label{eq:backward-loss}
\end{equation}

\subsubsection{Optimizing dialogue systems bidirectionally} Since the variable $z$ is sampled from the shared latent space between forward generation and backward reasoning steps, we can regard $z$ as a bridge to connect the training in two different direction and this opens the possibility to train dialogue models effectively. By merging Eq.~\ref{eq:forward-loss} and Eq~\ref{eq:backward-loss}, we can rewrite the lower bound Eq.~\ref{eq:loss-lb} as:
\begin{equation}
\begin{split}
\mbox{}\hspace*{-2mm}
\log {} &p(x,y \mid c) \geq \mathbb{E}_{z \sim q_\phi(z \mid c, x, y)} \left[
 \frac{1}{2} \log p_\theta (x \mid c, z, y)  \right. \\
& \phantom{XX}+ \frac{1}{2} \log p_\theta (y \mid c, z)  + \frac{1}{2}\log p_\theta (y \mid c, z, x) \\
& \phantom{XX}+ \left.\frac{1}{2} \log p_\theta (x \mid c, z) -D_\mathit{KL}(q_\phi(z \mid c, x, y) \| p_\theta(z \mid c))\vphantom{\frac{1}{2}}\right]\\
 ={}&L(c, x, y; \theta, \phi),
\end{split}
\label{eq:loss}
\end{equation}
%
which is the final loss function for our dialogue generation model. 

\subsubsection{Model architecture} The complete architecture of the proposed joint training process is shown in Figure~\ref{fig:framework}. 
It consists of three steps: (1) information encoding, (2) latent variable generation, and (3) target decoding. 
With respect to the information encoding step, we utilize a context encoder $Enc_{ctx}$ to compress the dialogue context $c$ while an utterance encoder $Enc_{utt}$ is used to compress the query $x$ and response $y$, respectively.
To model the latent variable $z$, we assume $z$ follows the multivariate normal distribution, the posterior network $q_\phi(z\mid c, x, y) \sim N(\mu, \sigma^2 I)$ and the prior network $p_\theta(z \mid c) \sim N(\mu^\prime, \sigma^{\prime2} I)$. Then, by applying the reparameterization trick \citep{kingma2013auto}, we can sample a latent variable $z$ from the estimated posterior distribution $N(\mu, \sigma^2 \bm{I})$. During testing, we use the prior distribution $N(\mu^\prime, \sigma^{\prime2} \bm{I})$ to generate the variable $z$. The KL-divergence distance is applied to encourage the approximated posterior $N(\mu, \sigma^2 \bm{I})$ to be close to the prior $N(\mu^\prime, \sigma^{\prime2} \bm{I})$. 
According to Eq.~\ref{eq:loss}, the decoding step in the right side of Figure~\ref{fig:framework} consists of four independent decoders, $Dec_1$, $Dec_2$, $Dec_3$, and $Dec_4$, corresponding to $\log p(y\mid c,z,x)$, $\log p(x\mid c,z)$, $\log p(x\mid c,z,y)$ and $\log p(y\mid c,z)$, respectively. Decoder $Dec_1$ is used to generate the final response during the testing stage. To make full use of the variable $z$, we attach it to the input of each decoding step. Since we have the shared latent vector $z$ as a bridge, training for the two directions is not independent, and updating one direction will definitely improve the other direction as well. In the end, both directions will contribute to the final dialogue generation process.

\section{Experimental Setup}
\label{sec:experiments}
\subsection{Datasets}
We use two datasets.
First, the MovieTriples dataset~\citep{serban2016hred} has been developed by expanding and preprocessing the Movie-Dic corpus~\citep{banchs2012movie} of film transcripts and each dialogue consists of 3 turns between two speakers. We regard the first turn as the dialogue context while the second and third one as the query and response, respectively. In the final dataset, there are around 166k dialogues in the training set, 21k in the validation set and 20k in the test set. In terms of the vocabulary table size, we set it to the top 20k most frequent words in the dataset.

Second, the DailyDialog dataset~\citep{li2017dailydialog} is a high-quality multi-turn dialogue dataset. We split the dialogues in the original dataset into shorter dialogues by every three turns as a new dialogue. The last turn is used as the target response and the first as the context and the third one as the query. After preprocessing, we have 65k, 6k, and 6k dialogs in the training, testing and validation sets, respectively. We limit the vocabulary table size to the top 20k most frequent words for the DailyDialog dataset.

\subsection{Baselines}
\begin{description}[leftmargin=\parindent,nosep]
\item[\textbf{Seq2SeqAtt}] This is a LSTM-based~\citep{hochreiter1997long} dialogue generation model with attention mechanism~\citep{bahdanau2014neural}.
\item[\textbf{HRED}] This method~\citep{serban2016hred} uses a hierarchical recurrent encoder-decoder to sequentially generate the tokens in the replies.
\item[\textbf{VHRED}] This extension of HRED incorporates a stochastic latent variable to explicitly model generative processes that possess multiple levels of variability~\citep{serban2017hierarchical}. This is also the model trained with Eq.~\ref{eq:loss-vhred}.
\item[\textbf{MMI}] This method first generates response candidates on a Seq2Seq model trained in the direction of context-to-target, $P(y\mid c,x)$, then re-ranks them using a separately trained Seq2Seq model in the direction of target-to-context, $P(x\mid y)$, to maximize the mutual information~\citep{li2015diversity}.
\item[\textbf{DC}] This method incorporates side information in the form of distributional constraints, including topic constraints and semantic constraints~\citep{baheti2018generating}.
\item[\textbf{DC-MMI}] This method is a combination of \textbf{MMI} and \textbf{DC}, where the decoding step takes into account mutual information together with the proposed distribution constraints in the method \textbf{DC}. 
\end{description}

\subsection{Training details}
We implement our model, \emph{Mirror}\footnote{Codebase: \url{https://github.com/cszmli/mirror-sigir}}, with PyTorch in the OpenNMT framework~\citep{opennmt}. The utterance encoder is a two-layer LSTM~\citep{hochreiter1997long} and the dimension is 1,000. The context encoder has the same architecture as the utterance encoder but the parameters are not shared. The four decoders have the same design but independent parameters, and each one is a two-layer LSTM with 1,000 dimensions. 
In terms of the dimension of the hidden vector $z$, we set it to $160$ for the DailyDialog dataset while $100$ for MovieTriples.  The word embedding size is $200$ for both datasets. We use Adam~\citep{kingma2014adam} as the optimizer. The initial learning rate is $0.001$ and learning rate decay is applied to stabilize the training process.

\subsection{Evaluation}
We conduct a human evaluation 
on Amazon MTurk guided by~\citep{li2019acute}.
For each two-way comparison of dialogue responses (against Mirror), we ask annotators to judge which of two responses is more appropriate given the context. 
For each method pair (Mirror, Baseline) and each dataset, we randomly sample $200$ dialogues from the test datasets; each pair of responses is annotated by $3$ annotators.

\section{Results and Analysis}
\label{sec:results}

In Table~\ref{Table:results}, we show performance comparisons between Mirror and other baselines on two different datasets. According to Table~\ref{Table:results}(top), it is somewhat unexpected to see that HRED can achieve such close performance compared to Mirror on DailyDialog, given its main architecture is a hierarchical encoder-decoder model. We randomly sample some dialogue pairs for which HRED outperforms Mirror to see why annotators prefer HRED over Mirror. For many of these cases, Mirror fails to generate appropriate responses, while HRED returns generic but still acceptable responses given the context. When we have the back reasoning step in Mirror, we expect that it will lead to more informative generations. Still, it also increases the risk of generating responses with incorrect syntax or relevant but inappropriate responses. A possible reason for the latter is that the backward reasoning step has dominated the joint training process, which can degenerate the forward generation performance. 

The performance gap between Mirror and all approaches (including HRED) is large on the DailyDialog dataset (see Table~\ref{Table:results}(bottom)). 
\begin{table}[t]
\caption{Human evaluation using the MovieTriple and DailyDialog datasets.} 
\label{Table:results}
  \centering
\begin{tabular}{l l*{4}{c}}
\toprule
& \textbf{Method pair} & Wins  & Losses &Ties  \\
\midrule
\parbox[t]{4mm}{\multirow{6}{*}{\rotatebox[origin=c]{90}{\em (a) MovieTriple~}}} 
& Mirror vs. Seq2SeqAttn  &0.53 &0.37 &0.10 \\
& Mirror vs. HRED  &0.41 &0.40 &0.19 \\
& Mirror vs. VHRED  &0.45 &0.38 &0.17 \\
& Mirror vs. MMI  &0.48 &0.42 &0.10 \\
& Mirror vs. DC  &0.50 &0.33 &0.17 \\
& Mirror vs. DC-MMI &0.39 &0.35 &0.26 \\
\midrule
\parbox[t]{4mm}{\multirow{6}{*}{\rotatebox[origin=c]{90}{\em (b) DailyDialog~}}} 
& Mirror vs. Seq2SeqAttn  &0.50 &0.26 &0.24 \\
& Mirror vs. HRED  &0.49 &0.32 &0.19 \\
& Mirror vs. VHRED  &0.48 &0.37 &0.15 \\
& Mirror vs. MMI  &0.40 &0.34 &0.26 \\
& Mirror vs. DC  &0.45 &0.38 &0.17 \\
& Mirror vs. DC-MMI &0.47 &0.35 &0.18 \\
\bottomrule
\end{tabular} 
 \vspace*{-0.5\baselineskip}
\end{table}
Due to space limitations, we only present one dialogue example in Table~\ref{Table:case_table}. The example is a typical case of why the response generated by DC has high embedding scores, but the human evaluation result is not promising. In this example, the response from DC has high semantic similarity with the context because of words like ``ask you'', ``apartment'', and ``questions''. However, it cannot be regarded as an appropriate and meaningful response in the given context. Comparing Mirror with methods that have use MMI (MMI, DC-MMI), the performance gap is relatively small. This is evidence showing the effectiveness of maximizing mutual information in improving the response quality. The Mirror method can be treated as a way to maximize mutual information implicitly. The advantage is that we can train dialogue models in two directions simultaneously.  

\begin{table}[!ht]
  \centering
\caption{Example generated responses by different models when the dialogue context is given.}
\label{Table:case_table}
\resizebox{0.8\linewidth}{!}{
  \begin{tabular}{c}
    \toprule
    \textbf{Context}\\
    \midrule
    \makecell[lt]{\textbf{Speaker A:} here ' s my license . \\
                   \textbf{Speaker B:} i ' m afraid i ' m going to have to ask you to\\ remain in the apartment . the narcotics squad will be arriving \\any moment now . they want to ask you a few questions . }  \\
    \midrule
    \textbf{Response} \\
    \midrule
    \makecell[lt]{\textbf{Reference:} squad ? what do they want with me ?\\ i don ' t even use aspirin !\\ 
                 \textbf{Seq2Seq:} no .\\
                \textbf{HRED:} i don ' t think so .\\ 
                \textbf{VHRED:} oh , i ' m sorry . \\ 
                \textbf{MMI:} i ' m sorry . i ' m sorry . i don ' t know what you ' re \\talking about . i don ' t know what i ' m afraid of . \\ 
                \textbf{DC:} i ' m not going to ask you . but he will be in the apartment for \\a moment -- and we can have some questions with that one of them !\\
                \textbf{DC-MMI:} i ' m going to ask you . \\
                \textbf{Mirror:} well , i ' m sure they ' ll have to wait . } \\

    \bottomrule
\end{tabular}
}
 \vspace*{-0.5\baselineskip}
\end{table}

\section{Conclusion and Future Work}
\label{sec:conclusion}
We have presented a novel approach to generating informative and coherent responses in open-domain dialogue systems, called \emph{Mirror}. First, we reformulate the original response generation task from two sides: context and response, to three sides: context, query, and response. Given the dialogue context and query, predicting the response is exactly like the traditional dialogue generation setup. Thus, \emph{Mirror} has one more step: inferring the query given the dialogue context and response.
By incorporating the backward reasoning step, we implicitly push the model to generate responses that have closer connections with the dialogue history. 
By conducting experiments on two datasets, we have demonstrated that Mirror improves the response quality compared to several competitive baselines without incorporating additional sources of information, which comes with additional computational costs and complexity. 
For future work, Mirror's bidirectional training approach can be generalized to other domains, such as task-oriented dialogue systems and question-answering tasks.

\clearpage

\bibliographystyle{ACM-Reference-Format}
\bibliography{bibliography}

\end{document}